\pdfoutput=1

\documentclass[11pt]{article}

\usepackage[]{ACL2023}

\usepackage{times}
\usepackage{latexsym}

\usepackage[T1]{fontenc}

\usepackage[utf8]{inputenc}

\usepackage{microtype}

\usepackage{inconsolata}

\usepackage{comment}
\usepackage{amsmath}
\usepackage{amssymb}
\usepackage{makecell}
\usepackage{multirow}
\usepackage{colortbl}
\usepackage{tipa}
\usepackage{float}
\usepackage{graphicx}
\definecolor{light}{rgb}{0.8,0.8,0.8}

\title{CUPE: Contextless Universal Phoneme Encoder for Language-Agnostic Speech Processing \footnotemark}

\author{Abdul Rehman \\
  \\
 \\
  \\\And
  Jian-Jun Zhang \\
  Bournemouth University \\
  Bournemouth, United Kingdom \\
  \texttt{arehman, jjunzhang, xyang@bournemouth.ac.uk} \\\And
  Xiaosong Yang \\
   \\ 
   \\
   \\}

\usepackage{float}
\begin{document}
\maketitle
\begin{abstract}
Universal phoneme recognition typically requires analyzing long speech segments and language-specific patterns. Many speech processing tasks require pure phoneme representations free from contextual influence, which motivated our development of CUPE - a lightweight model that captures key phoneme features in just 120 milliseconds, about one phoneme's length. CUPE processes short, fixed-width windows independently and, despite fewer parameters than current approaches, achieves competitive cross-lingual performance by learning fundamental acoustic patterns common to all languages. Our extensive evaluation through supervised and self-supervised training on diverse languages, including zero-shot tests on the UCLA Phonetic Corpus, demonstrates strong cross-lingual generalization and reveals that effective universal speech processing is possible through modeling basic acoustic patterns within phoneme-length windows. 
\end{abstract}

\section{Introduction}

\footnotetext{Accepted in: 8th International Conference  on Natural Language and Speech Processing  (ICNLSP 2025)} 

Current speech processing systems depend heavily on contextual information, creating a double-edged sword for certain tasks. While extensive context provides crucial bias toward appropriate attention mechanisms, it simultaneously makes it nearly impossible to isolate individual speech units—particularly allophones—from their contextual embeddings. Modern systems such as derivatives of wav2vec 2.0~\cite{baevski2020wav2vec} typically analyze 300-2500ms of speech, incorporating extensive language-specific patterns and contextual dependencies. While effective for automatic speech recognition, this approach entangles phonetic content with contextual information, making it extremely difficult to disentangle the acoustic properties that define individual speech sounds.

The necessity for contextless processing emerges from two critical considerations: alignment precision and representation purity. Extended temporal windows (e.g., 500ms) reduce inter-frame discriminability as individual frame representations become increasingly influenced by surrounding context. Optimal alignment performance requires maximally discriminative frame-level representations, where each frame maintains distinct characteristics. As context window length increases, the transformer's attention mechanism progressively attenuates frame-specific features through contextual averaging, resulting in diminished temporal resolution.

For paralinguistic tasks, contextless models function as quantization preprocessing stages. When frame-level embeddings encode predominantly contextual rather than local information, this homogenization undermines the model's capacity to capture subtle local acoustic variations essential for allophone analysis and speaker-specific phonetic characterization.

Our empirical results directly challenge the assumption that more context is always better—models using 120ms of speech windows actually perform on-par if not better than those using full word context across multiple evaluation scenarios, while simultaneously providing access to pure phonemic representations less contaminated by contextual dependencies.

Our work makes three key contributions. First, we demonstrate that universal phoneme recognition can be achieved effectively with just 120ms of context, a fraction of the 300-2500ms typically used in current approaches. Second, we introduce CUPE, a lightweight architecture (30M parameters) that achieves competitive performance through focused local feature extraction. Third, we provide a feature extraction method that captures pure phonemic representations by eliminating contextual dependencies, leading to cleaner and more interpretable phoneme embeddings across languages. By operating on brief windows—approximately the duration of a typical phoneme~\cite{crystal1988segmental}, CUPE learns language-agnostic acoustic features that characterize phonemes universally. This focus on fundamental acoustic patterns, independent of language-specific context, enables robust cross-lingual generalization and, crucially, provides access to clean allophonic representations that are essential for understanding speaker-specific phonetic variations.

The contextless nature of our approach enables several practical applications:
\begin{itemize}
    \item \textbf{Timestamps Alignment}: Generating time-aligned transcripts from raw text and audio. This task is critical for training downstream text-to-speech models. Since this is the main application for phoneme recognition, it helps to have as little context information in each frame so that there is a sharper contrast between frames for precise boundary detection.
    \item \textbf{Speech style learning}: It serves as a foundational allophone encoder. Embeddings of each frame can be used to generate acoustically pure allophone variants of base phonemes. This is also useful for training downstream text-to-speech tasks which currently rely on IPA dictionaries or sub-word tokens.
    \item \textbf{Robust phoneme verification}: Complementing traditional ASR systems by detecting and correcting errors that arise from over-reliance on language context.
    \item \textbf{Cross-linguistic research}: Generating language-agnostic phoneme representations that facilitate multilingual studies and enable more accurate speech disorder diagnostics.
\end{itemize}

Through extensive evaluation, we validate CUPE (Contextless Universal Phoneme Encoder), an architecture that deliberately restricts analysis to short windows. Our results demonstrate that this constrained approach matches or exceeds the performance of context-heavy models (XLS-R \cite{babu2022xlsr}) across diverse languages while using an order of magnitude fewer parameters and providing clean, context-independent phonemic representations suitable for allophone analysis.

\begin{figure}[ht!]
\centering
  \includegraphics[width=0.75\linewidth]{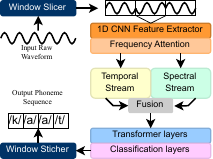}
  \caption{The windowing approach restricts the model's context for better localized learning, therefore, generalizing better across languages without learning longer patterns.}
  \label{fig:architecture}
\end{figure}

\section{Contextless Universal Phoneme Model}

Analysis of our evaluation datasets (Table~\ref{tab:datasets}) shows phoneme durations averaging 80ms (range: 62-107ms), consistent with Crystal and House's findings of 70-120ms for English phonemes \cite{crystal1988segmental}. Our architecture processes acoustic features through Conv1D layers at 13.1ms per frame, with a 120ms window and 80ms stride to capture 1-2 phonemes per window. This approach provides precise frame-level analysis while maintaining phoneme-level context, departing from traditional methods that rely on broader windows. To preserve acoustic continuity across overlapping windows, we implement a cosine-based weighting mechanism for feature fusion. The complete model architecture is illustrated in Figure \ref{fig:architecture}, with detailed specifications provided in Table \ref{tab:model_architecture}.

\subsection{Window Slicer}
The Window Slicer module addresses the fundamental challenge of processing continuous speech signals by segmenting raw waveforms (16\,kHz) into overlapping windows. This design enables localized feature extraction while preserving temporal continuity at boundaries. Using a 120\,ms window size with an 80\,ms stride provides sufficient context for phonetic events while reducing computational complexity from $\mathcal{O}(T^2)$ to $\mathcal{O}(W^2)$, where $T$ is the total sequence length and $W$ is the window size.

Given an input audio signal $\mathbf{x} \in \mathbb{R}^{B \times T}$, where $B$ is the batch size and $T$ is the total number of samples in the input sequence ($T = \text{sample\_rate} \times \text{duration}$):

\begin{equation}
    w_{b,i}(t) = x_b(t + is), \quad t \in [0, W-1]
\end{equation}

\noindent where $b \in [0, B-1]$ is the batch index, $i \in [0, N-1]$ is the window index, $t$ is the time index within each window, $W = 1920$ is the window size (120\,ms $\times$ 16\,kHz), $s = 1280$ is the stride length (80\,ms $\times$ 16\,kHz), and $N = \left\lfloor\frac{T-W}{s}\right\rfloor + 1$ is the number of windows.

\subsection{Feature Extractor}
Drawing from raw waveform processing techniques~\cite{dai2017rawCNN,schneider2019wav2vec}, our feature extraction stage implements a hierarchical CNN architecture that processes raw waveforms directly. This design, detailed in Table~\ref{tab:model_architecture}, captures increasingly abstract representations while maintaining computational efficiency. Following the success of Squeeze-and-Excitation Networks~\cite{hu2018squeeze} in speech recognition~\cite{han2020contextnet}, we incorporate adaptive channel-wise recalibration through frequency attention. The architecture separates temporal and spectral processing streams, inspired by multi-stream approaches~\cite{han2021multistream}, to capture both evolving acoustic patterns and frequency relationships.

\subsection{Windowwise Transformer}
Our transformer encoder layers process independent fixed windows instead of the whole clip, modifying the contextual processing of standard transformers~\cite{vaswani2017attention}. This approach represents a departure from traditional speech transformers by restricting context to local windows, ensuring that phoneme recognition decisions rely on relevant local context. Our preliminary experiments showed a tendency to overfit with larger transformer layers, leading us to maintain a light architecture (13M parameters for transformer) with a high dropout of 0.25. For comparison, the XLSR model~\cite{conneau2021unsupervised} has over 300M parameters.

\subsection{Classification and Window Stitching}
The final stage of our pipeline consists of classification and temporal integration. The transformer outputs first undergo classification through a two-layer neural network, which maps the high-dimensional representations to phoneme logits. This classifier is designed to untangle complex phonetic representations while maintaining computational efficiency.
To ensure temporal coherence across window boundaries, we implement a cosine-based weighting scheme:
\begin{equation}
\begin{split}
\tilde{y}(b,t,c) = \frac{\sum_{k} \cos(\pi t/F_w - \pi/2) \cdot y_k(b,t,c)}
{\sum_{k} \cos(\pi t/F_w - \pi/2) + \epsilon}, \\
t \in [0,F_w]
\end{split}
\end{equation}
where $y_k(b,t,c)$ represents the logit from window $k$ for batch $b$, time $t$, and class $c$. This weighted stitching approach enables effective recognition of phonemes shorter than the window length while preserving temporal coherence.

\begin{table*}[!ht]
\fontsize{8}{10}\selectfont
\centering
\caption{Detailed architecture specifications of the CUPE model with 30M trainable parameters.}
\begin{tabular}{l|cccccl}
\hline
Layer                                                                 & Output Shape & Parameters     & TR                    & RF                         & Other Details                                   \\ \hline
$\rightarrow$Window                                                                         & (B, 1, 1920) & -              & 80ms                  & 120-360ms                      & Speech waveforms at 16kHz                    \\
\arrayrulecolor{light}\hline Conv1D-1  & (B, n, 275)  & k=15, s=7, p=7 & 13.1ms                & 150ms                      & + BatchNorm + GELU + D(0.1)                     \\ \hline
Conv1D-2                                                             & (B, 2n, 55)  & k=11, s=5, p=5 & 1.9ms                 & 21.3ms                     & + BatchNorm + GELU + D(0.1)                     \\ \hline
Conv1D-3                                                            & (B, 4n, 19)  & k=7, s=3, p=3  & 0.4ms                 & 4.2ms                      & + BatchNorm + GELU + D(0.1)                     \\ \hline
Conv1D-4                                                                & (B, 8n, 10)  & k=5, s=2, p=2  & 0.1ms                 & 1.3ms                      & + BatchNorm + GELU + D(0.1)                     \\ \hline
Freq. Attention                                                        & (B, 8n, 10)  & k=1, s=1, p=0  & 0.1ms                 & 0.6ms                      & $\odot$ AvgPool+Conv1D+Sigmoid          \\ \hline
Temporal                                                                & (B, 8n, 10)  & k=7, s=1, p=3  & \multirow{2}{*}{13.1ms} & \multirow{2}{*}{±11 frames} & \multirow{2}{*}{2$\times$Conv1d, g=8, BN, GELU} \\
Stream (TS)                                                            & (B, 8n, 10)  & k=3, s=1, p=1  &                       &                            &                                                 \\ \hline
Spectral                                                               & (B, 12n, 10) & k=1, s=1, p=0  & \multirow{2}{*}{13.1ms} & \multirow{2}{*}{1 frame}   & \multirow{2}{*}{2$\times$Conv1d, g=8, BN, GELU} \\
Stream (SS)                                                            & (B, 8n, 10)  & k=1, s=1, p=0  &                       &                            &                                                 \\ \hline
Fusion                                                              & (B, 8n, 10)  & k=1, s=1, p=0  & 13.1ms                  & \multicolumn{2}{c}{Concat (TS, SS) + 1x1 Conv + BN + GELU}          \\ \hline
Transformer                                                          & (B, 10, 512) & $F_w$=10@120ms & 13.1ms                  & Full window                & 4 layers, 8 heads, Pre-norm, D(0.25)            \\ \hline
$\hookrightarrow$FT-Classifier                                                            & (B, 10, C)   & D=0.25              & 13.1ms                  & Full window                & Supervised only (512→2048→C)                  \\ \hline
$\hookrightarrow$PT-Projection                                                             & (B, 10, 256)   & D=0.25              & 13.1ms                  & Full window                & Unsuperv. only (512→2048→256)                  \\

\arrayrulecolor{black}\hline 
\multicolumn{6}{l}{TR: Temporal Resolution, RF: Receptive Field, B: batch size, k: kernel, s: stride, p: padding, n: base channels (256)} \\
\end{tabular}
\label{tab:model_architecture}
\end{table*}

\section{Experimentation}

We experiment with both supervised and self-supervised learning for the proposed model. First, we evaluated our model architecture using labeled speech and phoneme sequences. Then, we adapted the same architecture for self-supervised pretraining using vector quantization projections as targets, following a wav2vec-inspired approach. For baseline comparison, we use the XLS-R \cite{babu2022xlsr} 300M architecture with an additional linear classification layer. In non-pretrained evaluations, we reset XLS-R's parameters, while for pretrained evaluations, we fine-tune the off-the-shelf model with optional feature extraction layer freezing. The experimental pipeline remains consistent across all tests, varying only in context length (120ms, 160ms, 360ms, or complete words), model selection (XLS-R or CUPE), XLS-R parameter reset status, and feature extraction layer freezing status.

\subsection{Datasets}
\label{sec:datasets}

We evaluate our model on three diverse speech corpora:

\noindent\textbf{(1) FLEUR} (Few-shot Learning Evaluation of Universal Representations of Speech) \cite{conneau2023fleurs}: Used exclusively for self-supervised pretraining, comprising 5 hours of audio data from each of 102 languages. Table \ref{tab:datasets} reports trimmed durations excluding leading and trailing silences.

\noindent\textbf{(2) Multilingual Spoken Words Corpus (MSW)} \cite{mazumder2021mswc}: Contains isolated words from Mozilla Common Voice. We use 32 high-resource languages for training (10-hour limit per language) and 6 low-resource languages (lt, mt, ia, sk, ka, as) for evaluation. Twelve languages were excluded due to incompatibility with espeak-NG \cite{espeak_ng_2022}, the tool we used to generate IPA phoneme sequences from text.

\noindent\textbf{(3) UCLA Phonetic Corpus (UPC)} \cite{upc2021multilingual}: Features phonetically transcribed speech from 95 languages. We partition this dataset based on language overlap with XLS-R pretraining and FLEUR: UPC-eval contains 64 previously unseen languages, while UPC-seen includes 25 languages present in both pretrained XLS-R and FLEUR. The remaining six languages (fa, ig, kea, ab, eu, haw), exclusive to either XLS-R or FLEUR, serve as validation data during supervised training.

Table \ref{tab:datasets} summarizes the dataset statistics. The corpora differ significantly in language family distribution and recording conditions. MSWC and FLEUR predominantly feature Indo-European languages by duration, while UPC comprises 48.5\% African languages. MSWC offers diverse speakers and recording environments per language, whereas UPC contains just 60 utterances per language, typically from a single speaker in consistent recording conditions.

\subsection{Pre-Processing}

One of the fundamental challenges in creating a universal phoneme recognition system is accommodating unique phoneme inventories across languages. Prior work has explored two main approaches: probabilistic matching \cite{liu2023hierarchical, xinjian2021hierarchical}, which maps phonemes from new languages to acoustically similar training phonemes, and attribute-based decomposition \cite{glocker2023allophant}, which reconstructs language-specific phonemes from 35 articulatory attributes using the target language's IPA inventory. While both enable automated adaptation to new languages, they face tradeoffs in precision and feature completeness. Our approach instead employs systematic manual mapping of rare phonemes to standardized phoneme classes, prioritizing perceptual similarity over articulatory phonological relationships. Our mapping preserves high-frequency palatalized consonants (\textipa{t\super j}, \textipa{n\super j}, \textipa{r\super j}) while merging less frequent ones, maintains perceptually distinct vowel contrasts (e.g., \textipa{2} vs \textipa{@}, \textipa{I} vs \textipa{i}), keeps length distinctions for frequent vowels (\textipa{a:}, \textipa{e:}, \textipa{i:}, \textipa{o:}, \textipa{u:}), and maps rare phonemes to frequent counterparts based on confusion patterns (e.g., \textipa{6} → \textipa{a}, \textipa{C} → \textipa{k}). For affricates, we maintain distinct representations for common ones (\textipa{ts}, \textipa{tS}, \textipa{dZ}) while simplifying rare variants (\textipa{pf} → \textipa{f}), guided by both frequency and confusion patterns. The mapping dictionary is publicly available along with the source code to facilitate adoption and improvement.


\begin{table}[h]
\fontsize{8}{10}\selectfont
\centering
\caption{Datasets' details. $L_n$: total languages or lang code.  }
\begin{tabular}{l|ccccc}
\hline

Set     &  $L_n$  &  Hrs & WD(std)         & PPW(std)        & $U/C$        \\

\hline

MSWCtrain   & 32  & 181   & 0.80(0.12) & 6.30(1.45) & 803/65 \\
MSWCeval    & 6   & 15.6  & 0.82(0.12) & 6.39(1.34) & 117/56 \\
Lithuanian  & lt  & 5.2   & 0.87(0.12) & 6.58(1.34) & 66/42  \\
Maltese     & mt  & 4.9   & 0.77(0.11) & 6.25(1.30) & 56/38  \\
Interlingua & ia  & 3.17  & 0.84(0.12) & 5.98(1.26) & 29/29  \\
Slovak      & sk  & 1.37  & 0.88(0.11) & 6.77(1.3)  & 43/38  \\
Georgian    & ka  & 0.87  & 0.82(0.11) & 6.97(1.33) & 34/28  \\
Assamese    & as  & 0.05  & 0.77(0.12) & 5.80(1.21) & 31/26  \\
\arrayrulecolor{light}\hline
UPC-eval    & 67  & 0.82  & 0.93(0.20) & 5.01(1.53) & 237/59 \\
UPC-seen    & 28  & 0.56  & 0.89(0.22) & 4.89(1.32) & 221/60 \\
\arrayrulecolor{light}\hline
FLEURS      & 102 & 455   & -          & -          &     -      \\

\arrayrulecolor{black}\hline

\multicolumn{6}{l}{ WD: avg. Word Duration (s), PPW: avg. Phonemes-Per-Word } \\
\multicolumn{6}{l}{$U$: Unmapped unique phonemes,  $C$: Mapped phoneme classes.} \\
\end{tabular}
\label{tab:datasets}
\end{table}

\subsection{Supervised Training}

For each window, the model generates frame-level logits (10 frames per 120ms window, 28 frames for 360ms), which are stitched into continuous phoneme sequences. Training uses CTC loss \cite{graves2012connectionist} with an additional silence-awareness term:

\begin{equation}
\mathcal{L}1 = \mathcal{L}{\text{ctc}} + \alpha_s\mathcal{L}_{\text{sil}}
\end{equation}

\begin{equation}
\mathcal{L}_{\text{sil}} = \frac{1}{B}\sum_{t,b}(0.5\tilde{y}^t_b M^t_s + 0.1\tilde{y}^t_b(1-M^t_s))
\end{equation}

where $\tilde{y}^t_b$ is the blank token probability, $M^t_s$ is the silence mask, $B$ is batch size, and $\alpha_s$ (default 0.01) balances silence detection with phoneme recognition.

Our training pipeline optimizes for efficient learning through several mechanisms: AdamW optimizer with OneCycleLR scheduling, gradient norm clipping at threshold $\tau = 1.0$, and mixed-precision BF16 training for balanced efficiency and numerical stability. We trained all models on MSWC-train using a batch size (B) of 300 words until validation PER showed no further improvement, requiring 20 epochs and approximately 7 hours on two A6000 GPUs. The trained models and source code are available online\footnotemark[\value{footnote}], with results presented in Table \ref{tab:results_supervised}.

\subsection{Self-supervised Pre-Training}

For self-supervised pre-training, we modify CUPE by replacing the FT-Classifier with a prediction head (two projection layers with residual connections, layer normalization, GELU activation, and dropout 0.1) while being projected to a 256-dimensional feature space. The core architecture remains unchanged.

The pre-training uses masked prediction on 120ms windows (80ms stride), masking 40\% of features based on energy profiles and acoustic boundaries, with per-batch constraints of 10-80\%. A vector quantizer with 256-entry codebook serves as training target, using EMA updates (decay 0.99) and Laplace smoothing. The training objective combines reconstruction loss (smooth L1), contrastive loss with curriculum learning, codebook diversity loss, and similarity regularization.

Optimization uses AdamW (weight decay 0.05) with hierarchical learning rates (encoder: 5e-4, quantizer: 1e-3, prediction head: 1.5e-3) and one-cycle scheduling (15\% warmup, momentum 0.8-0.9). For evaluation, we freeze the feature extractor, replace the prediction head with classification layers, and fine-tune only the transformer and FT-Classifier components. We similarly evaluate XLS-R with both full and frozen-backbone fine-tuning.

\footnotetext{https://github.com/tabahi/contexless-phonemes-CUPE}

\begin{table*}[htb]
\fontsize{8}{10}\selectfont
\centering

\caption{Evaluation metrics (\%) for two architectures, XLSR (300M) \& CUPE (30M), trained on MSWC-train without pretraining.}
\begin{tabular}{l|cccc|cccccc|cccc} \hline
                      & \multicolumn{4}{c|}{Evaluation on   MSWC-eval}              & \multicolumn{6}{c|}{ Zero-shot PER on individual langs}                                          & \multicolumn{4}{c}{Zero-shot   evaluation on UPC-eval}             \\
Model:Context         & PER$\downarrow$           & GPm         & GPw           & F1           & lt            & mt          & ia            & sk            & ka            & as            & PER$\downarrow$           & GPm           & GPw           & F1           \\    \hline
XLSR:word & 49.9          & 35          & 51.7          & 60.6          & 59.5          & 48.7        & 37            & 45.3          & 48.4          & 65.8          & 66.5          & 31.2          & 51.7          & 52.9          \\
XLSR:120ms            & 52.6          & 34          & 52.1          & 59.9          & 61.1          & 49.9        & 42.9          & 52.3          & 50.4          & 63.7          & 66.3          & 31.6          & 51.1          & 54.9          \\ \arrayrulecolor{light}\hline
CUPE:word             & 46.4          & 39          & 55.1          & 63            & 54.5          & 47.1        & 33.1          & 42.5          & 44            & 60.5          & 58.8          & 32.9          & 52.5          & 58.3          \\
CUPE:360ms            & \textbf{44.8} & 38.3 & 56.5          & 62.6          & \textbf{53.8} & 45.2        & \textbf{30.8} & \textbf{39.7} & 42.5          & 60.9          & \textbf{52.2} & 34.7          & 53.1          & 61            \\
CUPE:160ms            & 47.8          & 36          & 55            & \textbf{64.8} & 57.2          & 46.2        & 36.2          & 45.2          & 44            & 60.7          & 57.5          & 32.9          & 54.1          & 58.8          \\
CUPE:120ms            & 45.9          & \textbf{40}        & \textbf{57.5} & 64.5          & 54.6          & \textbf{45} & 33.9          & 43.6          & \textbf{42.2} & \textbf{60.2} & 56.9          & \textbf{35.1} & \textbf{56.4} & \textbf{67.7}

\\ \arrayrulecolor{black}\hline
\end{tabular}

\label{tab:results_supervised}
\end{table*}

\begin{table*}[htb]
\fontsize{8}{10}\selectfont
\centering
\caption{Evaluation metrics (\%) for pre-trained models CUPE-PT (30M, pretrained on FLEURS), fine-tuned on MSWC-train, compared with XLSR (300M, off-the-shelf pretrained on 128 languages) with or without frozen backbone (FB) feature extractor. The top 4 rows show the results for contextless (120ms) models, the bottom 4 rows show results for word-context models for reference. Only the UPC-eval languages are unseen languages for zero-shot evaluation.}
\begin{tabular}{l|cccc|cccccc|cc|cc} 
\hline
\multirow{2}{*}{\textbf{Model:Context}} & \multicolumn{4}{c|}{\textbf{Eval. on MSWC-eval}} & \multicolumn{6}{c|}{\textbf{PER$\downarrow$ on individual langs (seen)}} & \multicolumn{2}{c|}{\textbf{UPC-eval}} & \multicolumn{2}{c}{\textbf{UPC-seen}} \\ 
\cline{2-15}
& \textbf{PER$\downarrow$} & \textbf{GPm} & \textbf{GPw} & \textbf{F1} & \textbf{lt} & \textbf{mt} & \textbf{ia} & \textbf{sk} & \textbf{ka} & \textbf{as} & \textbf{PER$\downarrow$} & \textbf{GPm} & \textbf{PER$\downarrow$} & \textbf{GPm} \\ 
\hline\hline
\multicolumn{15}{c}{\textbf{\textit{120ms Context Models}}} \\
\hline
FB-XLSR & 65.8 & 36.2 & \textbf{60.2} & 51.4 & 69.6 & 70.7 & 55.6 & 55.0 & 63.6 & 84.5 & 66.3 & \textbf{43.5} & 67.8 & 43.5 \\ 
FB-CUPE-PT & 49.8 & 34.9 & 53.0 & 60.5 & 59.3 & 47.5 & 38.5 & 48.5 & 44.6 & 61.4 & 66.5 & 35.4 & 69.7 & 38.2 \\
\hline
XLSR & 52.2 & 38.2 & 56.7 & 62.3 & 60.9 & 48.5 & 43.0 & 51.8 & 50.8 & 67.1 & 63.6 & 37.8 & 60.9 & 45.8 \\
CUPE-PT & \textbf{45.6} & \textbf{41.2} & 58.1 & \textbf{64.0} & \textbf{54.5} & \textbf{45.2} & \textbf{33.5} & \textbf{47.9} & \textbf{43.6} & \textbf{62.1} & \textbf{56.2} & 36.4 & \textbf{57.6} & \textbf{44.2} \\
\hline\hline
\multicolumn{15}{c}{\textbf{\textit{Word Context Models}}} \\
\hline
{\color{darkgray}FB-XLSR} & \textbf{{\color{darkgray}43.5}} & \textbf{{\color{darkgray}40.0}} & \textbf{{\color{darkgray}58.4}} & \textbf{{\color{darkgray}68.1}} & \textbf{{\color{darkgray}53.9}} & \textbf{{\color{darkgray}42.8}} & \textbf{{\color{darkgray}30.1}} & \textbf{{\color{darkgray}38.2}} & \textbf{{\color{darkgray}37.3}} & \textbf{{\color{darkgray}55.3}} & {\color{darkgray}66.9} & \textbf{{\color{darkgray}48.5}} & {\color{darkgray}70.3} & {\color{darkgray}43.4} \\
{\color{darkgray}FB-CUPEPT} & {\color{darkgray}70.4} & {\color{darkgray}1.9} & {\color{darkgray}29.6} & {\color{darkgray}54.3} & {\color{darkgray}73.6} & {\color{darkgray}65.1} & {\color{darkgray}71.2} & {\color{darkgray}77.1} & {\color{darkgray}70.5} & {\color{darkgray}62.5} & {\color{darkgray}69.0} & {\color{darkgray}3.2} & {\color{darkgray}73.2} & {\color{darkgray}2.7} \\
\hline
{\color{darkgray}XLSR} & {\color{darkgray}46.6} & {\color{darkgray}36.3} & {\color{darkgray}53.6} & {\color{darkgray}66.7} & {\color{darkgray}56.7} & {\color{darkgray}44.6} & {\color{darkgray}35.5} & {\color{darkgray}39.7} & {\color{darkgray}44.2} & {\color{darkgray}63.8} & \textbf{{\color{darkgray}46.9}} & {\color{darkgray}39.8} & \textbf{{\color{darkgray}46.0}} & \textbf{{\color{darkgray}46.4}} \\
{\color{darkgray}CUPE-PT} & {\color{darkgray}46.1} & {\color{darkgray}38.1} & {\color{darkgray}56.1} & {\color{darkgray}61.4} & {\color{darkgray}54.2} & {\color{darkgray}45.6} & {\color{darkgray}35.5} & {\color{darkgray}41.7} & {\color{darkgray}42.6} & {\color{darkgray}60.4} & {\color{darkgray}56.8} & {\color{darkgray}37.9} & {\color{darkgray}54.0} & {\color{darkgray}46.2} \\
\hline
\end{tabular}
\label{tab:results_unsupervised}
\end{table*}

\subsection{Results}

\subsubsection{Evaluation Metrics}
We decoded model outputs using Greedy Best-First Search and evaluated using Phoneme Error Rate (PER), Ground-truth Probability (\textbf{GP}), and F1-score. GP and F1 are computed after optimal alignment of true and predicted sequences, excluding insertions and deletions. While PER assigns a full penalty (+1) for any substitution, insertion, or deletion, it doesn't measure the near-misses. We introduce GP (\textbf{GPm} for macro, \textbf{GPw} for class-weighted) to better evaluate fine-grained phonemic distinctions like duration variants (\textipa{i}/\textipa{i:}) and vowel contrasts  (\textipa{\ae}/\textipa{a}) that are preserved in our approach rather than merged. GP measures the model's probability assignment to ground-truth classes at aligned time steps. It can be intuitively understood as the proximity to truth, or conversely, the inverse of the distance from truth. This proximity measure instead of PER is more important for judging the quality of embeddings for latent tasks. 

Detailed analysis of model behavior is provided in Appendix \ref{sec:appendixA}. The confusion matrix in Figure \ref{fig:confusion} shows that contextless recognition errors follow phonetically meaningful patterns, with confusions primarily occurring between acoustically similar sounds (e.g., front vowels, voiced/voiceless consonant pairs) rather than random misclassifications. The phoneme probability distributions over time (Figure \ref{fig:timeline}) illustrate CUPE's temporal resolution capabilities, showing distinct probability peaks corresponding to ground truth phonemes and smooth transitions between adjacent sounds.

\subsection{Key Insights and Limitations}

Looking at Table \ref{tab:results_supervised}, CUPE demonstrates remarkable cross-lingual generalization despite having a fraction of XLSR's parameters. While the 360ms model shows slightly better PER, this can be misleading due to class imbalances - it performs better on long and common vowels like /\textipa{a:}/ but struggles with short but rare phonemes, highlighting why GPm is a more balanced metric. Note that both 360ms and 120ms models have the same frame length of 16ms, the only difference is the context length. The significant performance difference in UPC evaluations, even when XLSR:120ms uses the same windowing pipeline, suggests that model's heavy size could be an overfitting liability.

\begin{table*}[!htb]

\fontsize{8}{10}\selectfont
\centering
\caption{Zero-shot PER comparison on UPC (UCLA Phonetic Corpus) with other works. Our CUPE:120ms results are fine-tuned on language splits matched to each baseline study for fair comparison, which differ from the UPC-eval/UPC-seen partitions in Tables 3-4. Direct performance comparison is limited due to different phoneme mapping systems. $L_n$ = number of unseen test languages (of 95).
}
\begin{tabular}{lcc|l}
\hline
\textbf{Study} & \textbf{$L_n$} & \textbf{$\downarrow$ PER (\%)} & \textbf{Phoneme Inventory Approach} \\
\hline
\cite{xinjian2021hierarchical} & 47 & 51.2 & Epitran+Allovera+Panphon \\
Ours & 47 & 46.1 & Systematic mapping to 65 classes  \\

\arrayrulecolor{light}\hline

\cite{liu2023hierarchical} & 10 & 64.7 & Direct UPC inventory \\
Ours & 10 & 44.1 & Systematic mapping to 65 classes  \\

\arrayrulecolor{light}\hline

\cite{xinjian2022phone} & 77 & 64.2 & Bayesian tree-based estimation \\
Ours & 77 & 48.6 & Systematic mapping to 65 classes  \\

\arrayrulecolor{light}\hline

\cite{glocker2023allophant} & 84 & 45.62 & 35 articulatory attribute system \\

Ours & 84 & 48.98 & Systematic mapping to 65 classes  \\
\arrayrulecolor{black}\hline

\end{tabular}
\label{tab:comparisons}
\end{table*}

Table \ref{tab:results_unsupervised} reveals that while XLSR with a frozen feature extractor achieves better overall metrics, CUPE maintains competitive performance under significant constraints. Notable observations include XLSR's degraded performance on UPC with frozen features and CUPE's sharp performance drop with word-context windows, perhaps due to having to learn more phonemes per window while most parameters are frozen. The completely unfrozen CUPE model's results mirror those in Table \ref{tab:results_supervised} even though the learning rate was set 10 times less for fine-tuning. The best contextless model, CUPE-PT:120ms, does not perform as well as pre-trained XLSR with full word context, indicating that additional context and parameters benefit large-scale pretraining. Nevertheless, CUPE's effectiveness with frozen feature extractors shows that essential phonetic information is learned by the feature extractor within brief temporal windows during pretraining.  Another sharp degradation is noticeable for CUPE-PT word context compared to 120ms; it is possibly due to 30M parameters being not enough for longer sequences (1000ms vs 120ms).

Our approach faces several limitations in its current form. The fixed 120 ms window presents inherent trade-offs in phoneme recognition: too long for short stop consonants and insufficient for capturing long phonemes fully. The model shows the best recall of stop consonants, but the worst recall of infrequent vowels. This issue is particularly evident in languages with contrastive length distinctions, where the model struggles to maintain consistent performance across different phoneme durations. 

The performance gap between supervised and pre-trained+fine-tuned results points to architectural limitations in both the projection mechanism and loss objectives. The current projection approach may not optimally preserve phonetic features during self-supervised learning, while the loss objectives could better reflect the hierarchical nature of phonemic contrasts. Additionally, the relatively modest size of the model (30M parameters) may limit its capacity to capture the full complexity of cross-linguistic phonetic variations. Additionally, our systematic mapping of rare phonemes, while practical, may obscure certain phonological contrasts. Although we achieve competitive results on the UCLA Phonetic Corpus, direct comparisons with methods such as Epitran \cite{xinjian2021hierarchical} and Allophant \cite{glocker2023allophant} are challenging due to fundamentally different phoneme inventory approaches.

While CUPE demonstrates strong performance in contextless phoneme recognition, several limitations warrant discussion. The model's varying performance across language families suggests potential biases in the feature extraction process that merit further investigation. Some languages with distinct phonological structures or phoneme inventories may require specialized preprocessing or architectural adaptations to achieve optimal performance. Additionally, the fixed 120ms window size, while effective across our evaluation datasets, may not be optimal for all languages or phonetic contexts—some phonemes naturally require longer or shorter temporal windows for accurate characterization.

Most importantly, this work establishes the foundation for more complex speech analysis systems. We have demonstrated how to extract clean embeddings for individual allophones—the next critical step is implementing a sentence-level speech style encoder that learns from these contextless allophone embeddings. Such a system would enable comprehensive analysis of speaker characteristics, accent patterns, and speaking styles while maintaining the interpretability and cross-linguistic generalizability that contextless representations provide.

 While our approach achieves competitive results on the UCLA Phonetic Corpus compared to existing methods listed in Table \ref{tab:comparisons}, these comparisons should be interpreted cautiously - each method uses fundamentally different phoneme inventory systems, from Epitran's probabilistic mappings  \cite{xinjian2021hierarchical} to Allophant's 35 articulatory attributes  \cite{glocker2023allophant}, making direct performance comparisons less meaningful. Our choice of 65 systematically mapped classes represents a different trade-off between granularity and generalization. The 65 class system is pragmatic implementation which can be expanded depending on the dataset. We selected 65 phonemes by empirically analyzing their occurrence across MSWC’s 50 languages, including only those that appeared at least 10,000 times. While phoneme mapping can further reduce the number of classes, our findings show that the impact on error rate is limited. For instance, when we applied broad phoneme group mapping to reduce the set to just 15 phonemes, the PER on MSWC-eval dropped from 0.45 to 0.40.


\section{Conclusion}

Through this work, we have demonstrated that effective universal phoneme recognition can be achieved using brief 120ms windows of speech input. Our CUPE model achieves competitive performance while requiring an order of magnitude fewer parameters than current approaches. The model's success in cross-lingual generalization validates our core finding that essential phonetic information can be captured through focused analysis of brief speech segments. These results provide compelling evidence that extensive temporal context is not a requirement for robust speech processing tasks. While our approach has some limitations, particularly with very long phonemes and limited phoneme inventory, it opens promising directions for lightweight, language-agnostic speech processing systems.  CUPE's effectiveness has significant implications for real-world applications, from low-latency speech recognition and ASR self-learning to speech pathology diagnostics. Our results indicate that future speech processing systems may benefit from focusing on fundamental acoustic patterns rather than extensive contextual dependencies.

\bibliography{acl2023}
\bibliographystyle{acl_natbib}

\clearpage

\appendix

\section{Confusion Heatmaps}
\label{sec:appendixA}

\makebox[\textwidth][c]{%
\includegraphics[width=0.99\textwidth]{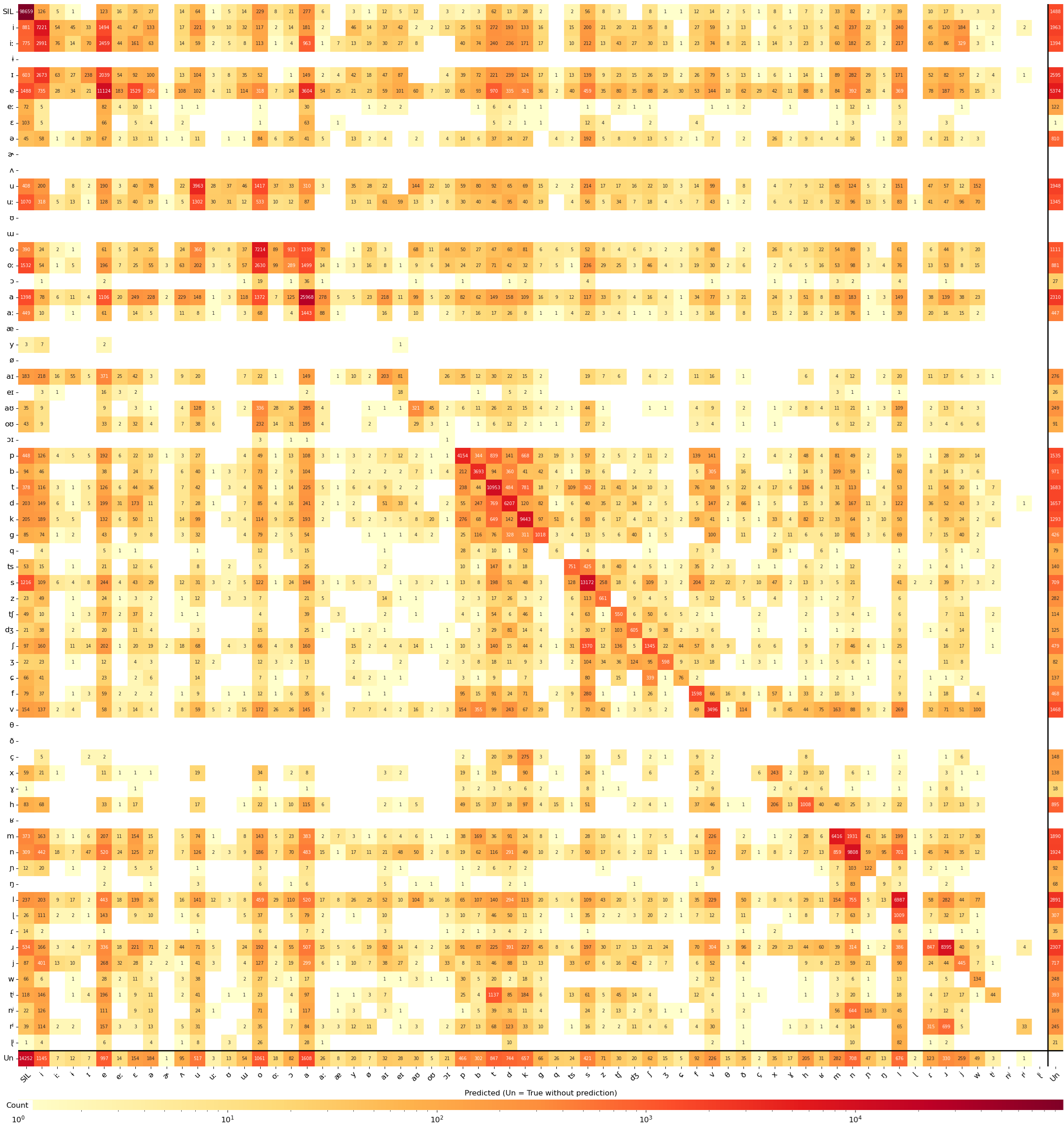}%
}

\makebox[\textwidth]{%
\parbox{0.95\textwidth}{%
\captionof{figure}{Confusion matrix for contextless phoneme recognition on MSWC-eval dataset using CUPE:120ms model trained on MSWC-train. The heatmap shows predicted phonemes (x-axis) versus ground truth phonemes (y-axis), with color intensity indicating count frequency on a logarithmic scale. The `Un' counts show the unaligned trues or predictions (i.e., the true sequence had a phoneme that didn't exist or aligned in the predicted sequence and vice-versa). The matrix reveals systematic confusion patterns, with darker cells along the diagonal indicating correct predictions. Notable off-diagonal clusters highlight acoustically similar phoneme pairs that are challenging for contextless recognition, such as front vowels, central vowels, and voiced/voiceless consonant pairs. The sparse structure demonstrates that most confusions occur within phonetically related categories rather than across distant phoneme classes.}
\label{fig:confusion}
}%
}

\begin{figure*}[!htb]
\centering
  \includegraphics[width=0.99\linewidth]{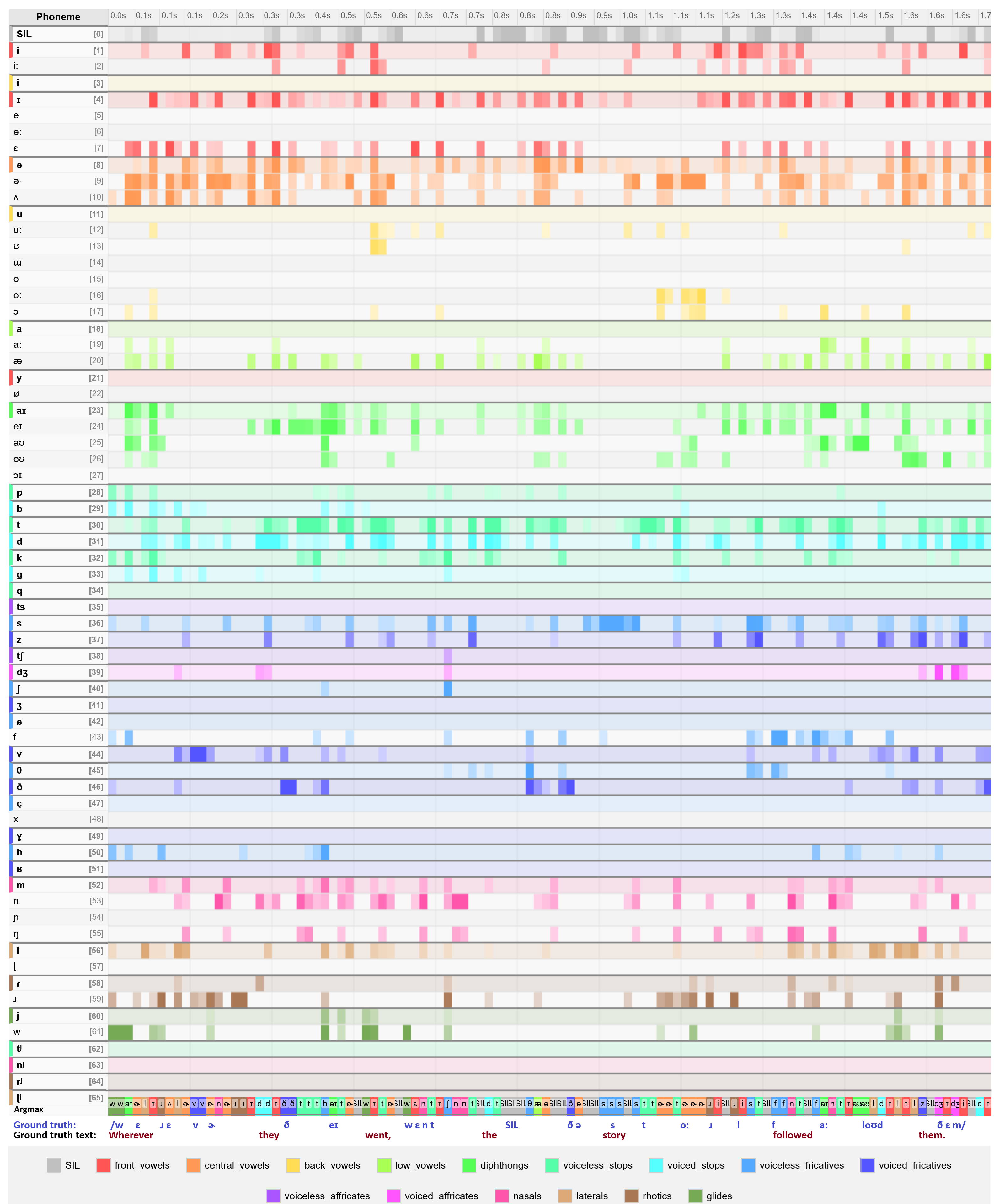}
  \caption{Phoneme probability distributions over time for an example utterance using CUPE:120ms model. The top panel shows a heatmap of phoneme probabilities (y-axis) across time frames (x-axis), with color intensity representing probability values. Ground truth phoneme alignments are displayed at the bottom with text. The visualization demonstrates the model's ability to capture temporal phoneme transitions in contextless recognition, with clear probability peaks corresponding to ground truth phonemes. Notable patterns include smooth transitions between phonemes within words and distinct silence regions (SIL) between words, highlighting the model's temporal resolution at 13ms.}
  \label{fig:timeline}
\end{figure*}

\end{document}